\documentclass[lettersize,journal]{IEEEtran}
\usepackage{amsmath,amsfonts}
\usepackage{algorithmic}
\usepackage{algorithm}
\usepackage{array}
\usepackage[caption=false,font=normalsize,labelfont=sf,textfont=sf]{subfig}
\usepackage{textcomp}
\usepackage{stfloats}
\usepackage{url}
\usepackage{verbatim}
\usepackage{graphicx}
\usepackage{cite}
\usepackage{multirow}
\usepackage{amsmath,amsfonts}
\usepackage{mathrsfs}
\usepackage{diagbox}
\usepackage{bm}
\usepackage{xspace}
\usepackage{color}

\newcommand{\Name}{\texttt{CDMA}\xspace}
\usepackage{enumitem}

\begin{document}
	
	\title{Boosting Black-box Attack to Deep Neural Networks with Conditional Diffusion Models}
	
	\author{
		Renyang Liu,
		Wei Zhou,~\IEEEmembership{Member,~IEEE,}
		Tianwei Zhang,~\IEEEmembership{Member,~IEEE,}
		Kangjie Chen,
		Jun Zhao,~\IEEEmembership{Member,~IEEE,}
		Kwok-Yan Lam,~\IEEEmembership{Member,~IEEE,}
		
		\IEEEcompsocitemizethanks{
			\IEEEcompsocthanksitem R. Liu is with the Information Science and Engineering School of Yunnan University, Kunming 650500, China (e-mail: ryliu@mail.ynu.edu.cn).
			\IEEEcompsocthanksitem W. Zhou is with the School of Software and the Engineering Research Center of Cyberspace, Yunnan University, Kunming 650500, China (e-mail: zwei@ynu.edu.cn).
			\IEEEcompsocthanksitem T. Zhang, K. Chen, J. Zhao and Kwok-Yan Lam are with the School of Computer Science and Engineering, Nanyang Technological University, Singapore (e-mail: tianwei.zhang@ntu.edu.sg; kangjie001@e.ntu.edu.sg; junzhao@ntu.edu.sg; kwokyan.lam@ntu.edu.sg).
		}
		
		
	}
	
	\markboth{Journal of \LaTeX\ Class Files,~Vol.~14, No.~8, August~2021}%
	{Shell \MakeLowercase{\textit{et al.}}: A Sample Article Using IEEEtran.cls for IEEE Journals}
	
	
	\maketitle
	
	\begin{abstract}
		Existing black-box attacks have demonstrated promising potential in creating adversarial examples (AE) to deceive deep learning models. Most of these attacks need to handle a vast optimization space and require a large number of queries, hence exhibiting limited practical impacts in real-world scenarios. In this paper, we propose a novel black-box attack strategy, Conditional Diffusion Model Attack (\Name), to improve the query efficiency of generating AEs under query-limited situations. The key insight of \Name is to formulate the task of AE synthesis as a distribution transformation problem, i.e., benign examples and their corresponding AEs can be regarded as coming from two distinctive distributions and can transform from each other with a particular converter. Unlike the conventional \textit{query-and-optimization} approach, we generate eligible AEs with direct conditional transform using the aforementioned data converter, which can significantly reduce the number of queries needed. \Name adopts the conditional Denoising Diffusion Probabilistic Model as the converter, which can learn the transformation from clean samples to AEs, and ensure the smooth development of perturbed noise resistant to various defense strategies. We demonstrate the effectiveness and efficiency of \Name by comparing it with nine state-of-the-art black-box attacks across three benchmark datasets. On average, \Name can reduce the query count to a handful of times; in most cases, the query count is only ONE. We also show that \Name can obtain $>99\%$ attack success rate for untarget attacks over all datasets and targeted attack over CIFAR-10 with the noise budget of $\epsilon=16$.
	\end{abstract}
	
	
	\section{Introduction}
	\label{sec:intro}
	In recent years, Deep Learning (DL) has experienced rapid development, and DL models are widely deployed in many real-world applications, such as facial recognition \cite{pr/ChenYTX22}, autonomous driving \cite{eccv/LiZOQ22}, financial services \cite{isci/ZhangHXW21}, etc. However, existing DL models have been proven to be fragile that they can be easily fooled by adding elaborately calculated imperceptible perturbations to the benign inputs, known as adversarial examples (AEs). Therefore, the security of DL models has been attracting more and more attention from researchers.
	
	Typically, adversarial attacks can be classified into two categories based on their settings. The first one is white-box attacks \cite{sp/Carlini017, make/CombeyLFH20}, where the attacker has complete information of the victim models, including the model structure, weights, gradients, etc. Such information can assist the attacker to achieve a very high attack success rate. A variety of attack techniques have been proposed to effectively generate AEs under the white-box setting, e.g., FGSM \cite{corr/GoodfellowSS14}, C\&W \cite{sp/Carlini017}, etc. 
	
	The second one is black-box attacks \cite{kdd/ChenG20,nips/DolatabadiEL20, kdd/ShuklaSWK21,eccv/WangZTGHLL22,pr/BaiWZJX23}, which is more practical in the real world. The attacker is not aware of the victim model's information. He has to repeatedly query the victim model with carefully crafted inputs and adjust the perturbations based on the returned soft labels (prediction probability) or even hard labels \cite{iclr/BrendelRB18,iclr/IlyasEM19}.
	Many query-efficient and transfer-based attack methods have been proposed recently. However, they suffer from several limitations. First, these methods still need hundreds to thousands of queries to generate one AE \cite{nips/DolatabadiEL20}), especially in the targeted attack setting. This makes the attack costly in terms of computation resources, time and monetary expense, restricting their practicality in real-world scenarios. Besides, more queries can remarkably increase the risk of being detected \cite{mm/ShahRKR21}. Second, AEs generated by the noise-adding manner are easy to be identified or denoised, decreasing the attack performance to a large extent \cite{icml/NieGHXVA22,corr/abs-2205-14969}. Once the victim model is equipped with some defense mechanisms, the attacker needs to consume more model queries to optimize a new AE. Third, the quality of the generated AE highly depends on the similarity between the local surrogate model and the victim model, which normally cannot be guaranteed. This also limits the performance of existing attack methods. 

	Driven by the above drawbacks, the goal of this paper is to design new hard-label black-box attack approaches, which can generate AEs with limited queries for both untarget and targeted settings. This is challenging due to the restrictions of limited information about the victim model, and query budget. Our observation is that \textit{clean samples and their corresponding AEs follow two adjacent distributions, connected by certain relationships}. This presents an opportunity to \textit{build a converter, which can easily transfer each clean sample to its corresponding AE without complex optimization operations}. 
	Following this hypothesis, we propose \Name, a novel Conditional Diffusion Model Attack to attack black-box DL models efficiently. Different from prior attacks using the iterative query-and-optimization strategy, \Name converts the AE generation task into an image translation task, and adopts a conditional diffusion model (i.e., the converter) to directly synthesize high-quality AEs. 
	In detail, we first execute the diffusion process to train a conditional diffusion model with pre-collected pair-wised clean-adversarial samples, where the AEs are generated with white-box attack methods from local shadow models. During the training, the clean images are used as the condition to guide the diffusion model to generate eligible AEs from a given unique input. Once the diffusion model is trained, we can execute the reverse process for the clean input to formulate corresponding AEs. 
	
	Compared to existing works, \Name has the following advantages. (1) It significantly improves the attack effectiveness by conditional synthesis instead of query and optimization. (2) \Name does not rely on the inherent attribute of the target model. It only requires the hard labels to verify whether the victim model has been attacked successfully. As a result, the pre-trained diffusion model has a high generalization ability to attack any DL models. (3) Once the diffusion model is well-trained, the attacker can batch-wisely sample sufficient candidate AEs, further improving the attack efficiency and scalability. (4) Benefiting from the smooth synthesizing processes, the formulated AEs are challenging to be purified and can keep high robustness against different defense mechanisms.
	
	We evaluate \Name on mainstream datasets (CIFAR-10, CIFAR-100 and Tiny-ImageNet-200), and compare it with state-of-the-art black-box attack methods, including pure black-box attacks (soft- and hard-label), query-based and transfer-based attacks. Extensive experiment results demonstrate our superior query efficiency. In all attack settings, \Name achieves a comparable attack success rate to all baselines but with significantly reduced numbers of queries. Besides, AEs generated from \Name exhibit higher robustness to several mainstream defense strategies. Finally, the empirical results of data-independent and model-independent attacks have validated our assumption, i.e., the clean and adversarial examples come from two disparate distributions, which can be transformed into each other, and the proposed \Name can well learn this transformation relationship.

	To summarize, our main contributions are as follows:
	\begin{itemize}[leftmargin=*, itemsep=1pt, topsep=1pt, parsep=0pt]
		\item We model the adversarial example generation as a distribution transform problem with a perfect data converter on certain conditions to achieve efficient black-box attacks.
		
		\item We build the data converter with a diffusion model and propose a novel diffusion model-based black-box attack named \Name, which can directly formulate the corresponding AE by conditional sampling on the original clean image without the complex iterative process of query and optimization.
		
		\item \Name can generate AEs with high attack ability and robustness. These AEs can be well transferred to different victim models and datasets.
		
		\item We perform extensive experiments to demonstrate the superiority of \Name over state-of-the-art black-box methods, in terms of query efficiency, attack robustness and effectiveness in both untargeted and targeted settings. 
	\end{itemize}

	The remainder of this paper is organized as follows: we briefly review the existing literature on adversarial attacks in Sec. \ref{Sec:related}. We define our distribution transformation-based attack and propose the diffusion model-based \Name method in Sec. \ref{sec:methodology}. In Sec. \ref{Sec:experiments}, we perform extensive experiments to show that \Name is more efficient and effective than other baseline attacks under untarget and targeted situations. It can also successfully keep the high attack performance against different defense strategies. Finally, we conclude this paper in Sec. \ref{Sec:conclusion}.

	\section{Related Work}
	\label{Sec:related}
	
	Adversarial attacks against deep learning models refer to the process of intentionally manipulating benign inputs to fool well-trained models. Based on the setting, existing attacks can be classified into two categories: in the \textit{white-box} setting, the attacker knows every detail about the victim model, based on which he creates the corresponding AEs. In the \textit{black-box} setting, the attacker does not have the knowledge of the victim model, and are only allowed to query the model for AE generation. In this paper, we focus on the black-box one, which is more practical but also more challenging.
	
	There are three types of techniques to achieve black-box adversarial attacks. The first one is transfer-based attacks. Papernot et al. \cite{ccs/PapernotMGJCS17} proposed the pioneering work towards black-box attacks, which first utilizes Jacobian-based Dataset Augmentation to train a substitute model by iteratively querying the oracle model, and then attacking the oracle using the transferability of AEs generated from the substitute model. TREMBA \cite{iclr/Huang020} trains a perturbation generator and traverses over the low-dimensional latent space. ODS \cite{nips/Tashiro0E20} optimizes in the logit space to diversify perturbations for the output space. GFCS \cite{iclr/LordMB22} searches along the direction of surrogate gradients and falls back to ODS. CG-Attack \cite{cvpr/FengWFL0X22} combines a well-trained c-glow model and CAM-ES to extend attacks. However, these transfer-based attacks heavily rely on the similarity between the substitute model and the oracle model.
	

	The second type is score-based attacks. Ilyas et al. \cite{iclr/IlyasEM19} proposed a bandit optimization-based algorithm to integrate priors, such as gradient priors, to reduce the query counts and improve the attack success rate. Chen et al. \cite{ccs/ChenZSYH17} proposed zeroth order optimization-based attacks (ZOO) to directly estimate the gradients of the target DNN for generating AEs. Although this attack achieves a comparable attack success rate, its coordinate-wise gradient estimation requires excessive evaluations of the target model and is hence not query-efficient. AdvFlow \cite{nips/DolatabadiEL20} combines a normalized flow model and NES to search the adversarial perturbations in the latent space to balance the query counts and distortion and accelerate the attack process.
	
	
	The third type is decision-based attacks, which are specifically designed for the hard-label setting. Boundary attack \cite{iclr/BrendelRB18} is the earliest one that starts from a large adversarial perturbation and then seeks to reduce the magnitude of perturbation while keeping it adversarial. Bayes\_Attack \cite{kdd/ShuklaSWK21} uses Bayesian optimization to find adversarial perturbations in the low-dimension subspace and maps it back to the original input space to obtain the final perturbation. NPAttack \cite{pr/BaiWZJX23} considers the structure information of pixels in one image rather than individual pixels during the attack with the help of a pre-trained Neural Process model. Rays \cite{kdd/ChenG20} introduces a Ray Searching Method to reformulate the continuous problem of finding the nearest decision boundary as a discrete problem that does not require any zero-order gradient estimation, which significantly improves the previous decision-based attacks. Triangle Attack (TA) \cite{eccv/WangZTGHLL22} optimizes the perturbation in the low-frequency space by utilizing geometric information for effective dimensionality reduction.

	The above query-and-optimization black-box attacks are inefficient and uneconomical because they require thousands of queries on the target model. In this situation, the time and computational consumption could be very considerable. On the other hand, the performance of transfer-based black-box attacks is often limited by the similarity between the surrogate model and the oracle model. Besides, these attacks cannot extend to the data-independent or model-independent scenario or keep robustness to different defense strategies, which fades the attack capability to a considerable extent.
	
	Therefore, it is necessary to have a method that can efficiently generate AEs within limited queries, which are effective against different models and datasets. We propose to use the diffusion model to achieve this goal. The diffusion model is an advanced technique for image translation tasks. We can train such a model to convert clean images to AEs against the black-box victim model. Our attack, \Name, does not require a large number of queries or detailed information regarding the victim model in the attacking process and the formulated AEs can be resistant to most defense strategies.
	
	\begin{figure*}[!htp]
		\centering
		\setlength{\abovecaptionskip}{0pt}
		\setlength{\belowcaptionskip}{-15pt}
		
		\includegraphics[width=\textwidth]{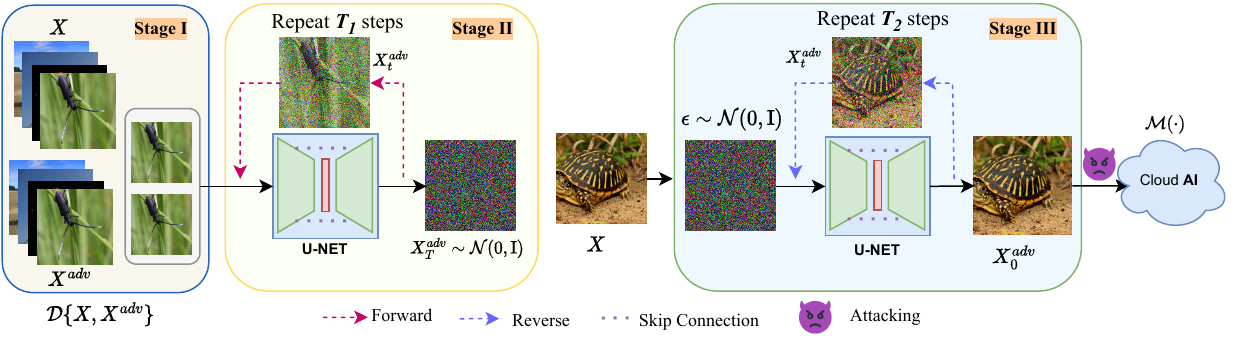}
		\caption{Overview of \Name. $\mathcal{D}\{X,X^{adv}\}$ is the collected pair-wised clean and adversarial dataset and $X_{t}^{adv}$ is the adversarial example $X^{adv}$ at the forward or reverse step $t$. $\epsilon \sim \mathcal{N}(0, \mathrm{I})$ is the Gaussian noise and $\mathcal{M}(\cdot)$ is the target victim model.}
		\label{fig:framework}
	\end{figure*}
	
	\section{Methodology}
	\label{sec:methodology}

	\subsection{Problem Definition}
	Given a well-trained DNN model $ \bm{\mathcal{M}} $ and an input $ \bm{x} $ with its corresponding label $ y $, we have $ \bm{\mathcal{M}}(\bm{x})=y $. The AE $ \bm{x}^{adv} $ is a neighbor of $ \bm{x} $ that satisfies $ \bm{\mathcal{M}}(\bm{x}^{adv}) \neq y $ and $ \left \| \bm{x}^{adv}-\bm{x} \right \|_p \leq \epsilon  $, where $ \bm{L}_p $ norm is used as the metric function and $ \epsilon $ is a small noise budget. With this definition, the problem of generating an AE becomes a constrained optimization problem:
	\begin{equation}
		\label{eq:eq1}
		\bm{x}_{adv}=
		\underset{\left \| \bm{x}^{adv}-\bm{x} \right \|_p \leq \epsilon}{\mathop{arg \ max}\mathcal{L}} ( \bm{\mathcal{M}}(\bm{x}^{adv}) \neq y),
	\end{equation}
	where $ \mathcal{L} $ stands for a loss function that measures the confidence of the model outputs.
	
	Existing attack methods normally utilize the information (e.g., the prediction results, model weights, etc.) obtained from the target model to optimize the above loss function. Different from them, in this paper, we convert the AE generation problem into an image-to-image task: an adversarial image $\bm{x}_{adv}$ can be regarded as a particular transformation from its corresponding clean image $\bm{x}$. These two different images ($\bm{x}$ and $\bm{x}_{adv}$) can be mutually transformed from each other by a converter. We choose a rising star generative model, the diffusion model, as our image converter and propose a Conditional Diffusion Model-based Attack framework for synthesizing AEs. 
	
	\subsection{Denoising Diffusion Probabilistic Models}
	Unlike VAE or Flow models, diffusion models are inspired by non-equilibrium thermodynamics to learn through a fixed process. The latent space has a relatively high dimensionality. It first defines a Markov chain of diffusion steps and corrupts the training data by continuously adding Gaussian noise until it becomes pure Gaussian noise. Then it reverses the process by removing noise and reconstructing the desired data. Once the model is well-trained, it can generate data through the learned denoising process by inputting randomly sampled noise. 
	Here, we briefly review the representative Denoising Diffusion Probabilistic Models (DDPM) \cite{nips/HoJA20}.
	
	In the forward progress (i.e., adding noise), given an image $x_0 \sim q(x)$, the diffusion process can obtain $x_1, x_2,...,x_T$ by adding Gaussian noise $T$ times, respectively. This process can be expressed as a Markov chain:
	\begin{equation}
		\begin{aligned}
			q(x_t|x_{t-1}) & = \mathcal{N}(x_t;\sqrt{1-\beta _t} x_{t-1},\beta _t \mathrm {I}),                                  \\
			q(x_{1:T}|x_0) & =\prod_{t=1}^{T}q(x_t|x_{t-1})= \prod_{t=1}^{T}(x_t;\sqrt{1-\beta _t}x_{t-1},\beta _t \mathrm {I} )
		\end{aligned}
	\end{equation}
	where $ t \in {1,2,...,T} $, $\left \{ \beta _t \in (0,1) \right \} _{t=1}^{T} $ is the hyper-parameter of the Gaussian distribution's variance. In this process, $x_t$ tends to be pure Gaussian noise with the increase of $t$. It finally becomes the Standard Gaussian noise $\mathcal{N}(0,\mathrm{I})$ when $T \to \infty $.
	
	Suppose $\alpha _t := 1- \beta _t$ and $\bar{\alpha }_t := {\textstyle \prod_{i=1}^{T} \alpha _i} $. Then $x$ of arbitrary $t$ can be written in the following closed form:
	\begin{equation}
		\begin{aligned}
			& q(x_t|x_0)  = \mathcal{N}(x_t;\sqrt{ \bar{a}_t } x_0,(1-a_t) \mathrm {I}), \\
			& x_t = \sqrt{\bar{\alpha} _t} x_0 + \sqrt{1-\bar{\alpha}_t} \delta
		\end{aligned}
	\end{equation}
	where  $ \delta \sim \mathcal{N}(0, \mathrm{I}) $. $x_t$ satisfies $q(x_t|x_0)=\mathcal{N}(x_t;\sqrt[]{\bar{a_t} }x_0,(1-\bar{a_t} )\mathrm {I}) $.
	
	The reverse process is the denoising of diffusion. If we can gradually obtain the reversed distribution $q(x_{t-1}|x_t)$, we can restore the original image $x_0$ from the standard Gaussian distribution $\mathcal{N}(0,\mathrm{I})$.
	
	As $ q(x_t|x_{t-1})$ is a Gaussian distribution and $\beta _t$ is small enough, $q(x_{t-1}|x_t)$ is a Gaussian distribution. However, we do not have a simple way to infer $q(x_{t-1}|x_t)$. DDPM adopts a deep neural network, typically U-Net, to predict the mean and covariance of $x_{t-1}$ of the given input $x_t$. In this situation, the reverse process can be written as the parameterized Gaussian transitions:
	\begin{equation}
		\begin{aligned}
			p_\theta (X_{0:T) = P(x_T})\prod_{t=1}^{T} p_\theta (x_{t-1}|x_t), \\ P_ \theta (x_{t-1}|x_t) = \mathcal{N}(x_{t-1};\mu(x_t,t), {\textstyle \sum_{\theta }^{}(x_t,t)} )
		\end{aligned}
		\label{eq:ddpm_sample_1}
	\end{equation}
	
	With Bayes's theorem, DDPM predicts the noise $\delta _\theta(x_t,t)$ instead and computes $\mu(x_t,t)$ as follows:
	
	\begin{equation}
		\mu(x_t,t)=\frac{1}{\sqrt{\alpha _t}} (x_t-\frac{\beta _t}{\sqrt{1-\bar{\alpha } _t}}\delta _\theta (x_t,t) )
	\end{equation}
	
	\subsection{Conditional Diffusion Model Attack (\Name)}
	The whole framework of \Name is illustrated in Figure~\ref{fig:framework}, which can be split into the following three stages: training sample collection, model training (forward process), and AE generating (reverse process). Specifically, in \textbf{Stage I}, the attacker collects the clean-adversarial example pairs, where the adversarial examples are built from local shadow models using standard white-box attack techniques. In \textbf{Stage II}, the attacker trains a conditional diffusion model with the pair-wised $(\bm{x}, \bm{x}_{adv})$ sampled from the pre-collocated dataset $\mathcal{D}\{X,X^{adv}\}$. The conditional diffusion model is composed of a series of encoder-decoder-like neural networks (UNET \cite{miccai/RonnebergerFB15} is adopted in this work). Once the model is well-fitted, the attacker can perform the attacks against the victim model in a sampling manner in \textbf{Stage III} instead of a query-and-optimization way. Below we give details of each stage.
	
	\subsubsection{\textbf{Training Sample Collection}}
	Recall that our training data are paired with clean and adversarial samples, where the clean example is used as an inference image and concatenated with its corresponding adversarial example to compose the diffusion model's input. More specifically, for a given dataset, we first use typical white-box attack methods (e.g., PGD \cite{iclr/MadryMSTV18} and et al.) to attack the local shadow model and obtain the corresponding adversarial examples, which are then paired with the original clean examples to formulate the training dataset $\mathcal{D}=\{X, X^{adv}\}$ of our diffusion model.
	
	\subsubsection{\textbf{Conditional Diffusion Model Training}}
	The core of training a diffusion model is to make it predict reliable noise $\delta$. Unlike \cite{nips/HoJA20}, we need to consider the additional conditional variable $x$. We use $\delta$ to represent the real noise added to $x^{adv}$ at each step $t$, and use $ \hat{\delta}_{\theta} $ to represent the noise predicted by model $f(\cdot)$ (U-NET in this paper). Then the final objective function can be written as:
	\begin{equation}
		\mathcal{L}=E_{t,\{x, x_{0}^{adv}\}, \delta}\left \| \delta - \hat{\delta}_{\theta} (x_t^{adv},t,x)  \right \|_{p}
	\end{equation}
	where $ t\sim[1,2,...,T],\{ x,x_{0}^{adv}\} \sim \mathcal{D}\{x,x^{adv}\}, x_{t}^{adv} \sim q(x_{t}^{adv}|x_{0}^{adv},x) $,  $\delta \sim \mathcal{N}(0,\mathrm{I})$, $\| \cdot \|_{p}$ represents the $L_p$-norm and $p\in \{0,1,2,L_{\infty}\}$. As demonstrated in \cite{siggraph/SahariaCCLHSF022}, $L_1$ yields significantly lower sample diversity compared to $L_2$. Since we aim to generate diversified adversarial examples, we also adopt $L_2$, i.e., MSE, as our loss function to constrain the true noise $\delta$ and the predicted noise $ \hat{\delta}_{\theta} $.
	
	\begin{algorithm}[!ht]
		\renewcommand{\algorithmicrequire}{\textbf{Input:}}
		\renewcommand{\algorithmicensure}{\textbf{Output:}}
		\caption{Conditional Diffusion Model Training}
		\label{alg:alg1}
		\begin{algorithmic}[1]
			\REQUIRE $ \{x,x^{adv}\} $: the clean image and adversarial image pair; $t \sim \mathcal{U} ({1,...,T})$: The time-steps belong to Uniform distribution.
			\ENSURE The well-trained model $M(\cdot)$.
			\REPEAT
			\STATE Take the gradient step on \\ $\mathcal{L}=E_{t,\{x,x_{0}^{adv}\},\delta}\left \| \delta - \hat{\delta}_{\theta} (x_t^{adv},t,x)  \right \|_{p} $.
			\UNTIL{converged}
		\end{algorithmic}
	\end{algorithm}
	
	\begin{algorithm}[!ht]
		\renewcommand{\algorithmicrequire}{\textbf{Input:}}
		\renewcommand{\algorithmicensure}{\textbf{Output:}}
		\caption{Conditional Diffusion Model Attacking}
		\label{alg:alg2}
		\begin{algorithmic}[1]
			\REQUIRE $ \bm{\mathcal{C}} $: the target model to be attacked;
			$ \bm{x} $: the clean image, the conditioning information for conditional sampling;
			$ \bm{Q} $: the maximum querying number;
			$ q $: the current querying number;
			$ \epsilon $: the noise budget.\\
			\ENSURE The adversarial example $\bm{x}_{adv}$ used for attack.
			\STATE $\bm{x}_{T}^{adv} \sim \mathcal{N}(\bm{0},\bm{I})$.
			\WHILE{$q<=Q$}
			\FOR{$t=T,...,1$}
			\STATE $\bm{z} \sim \mathcal{N}(\bm{0},\bm{I})$ if $t>1$, else $\bm{z}=0$
			\STATE $\bm{x_{t-1}} \sim q(x_{t-1}|x)$
			\STATE $\bm{x}_{t-1}^{adv}  \gets  p_{\theta}(\bm{x}_{t-1}^{adv}|\bm{x}_{t}^{adv},\bm{x})$
			\ENDFOR
			\STATE $\delta = clip(\bm{x}_{0}^{adv}-x,-\epsilon,\epsilon)$
			\STATE $ \bm{x}^{adv}=clip(x+\delta,0,1)$
			\IF{$ \bm{x}^{adv} $ attack $\bm{\mathcal{C}}$ successfully}
			\STATE break.
			\ENDIF
			\ENDWHILE
			\STATE \textbf{return} $ \bm{x}^{adv} $
		\end{algorithmic}
	\end{algorithm}

	\subsubsection{\textbf{Generate Adversarial Examples.}}
	In \Name, the attacker generates adversarial examples for benign images by sampling from the well-trained conditional diffusion model. Our generation process becomes sampling from the conditional distribution $P(x_{0}^{adv}|c)$, where $c$ is the clean image $x$. As the aforementioned sampling process of DDPM \cite{nips/HoJA20, iccv/ChoiKJGY21} (Eq. \ref{eq:ddpm_sample_1}), the conditional sampling can be written as follows:
	
	\begin{equation}
		\begin{aligned}
			p _\theta(x_{0}^{adv}|x) = \int p _\theta(x_{0:T}^{adv}|x) dx_{1:T}^{adv} \\
			p _\theta(x_{0}^{adv}|x) = p(x_{T}^{adv})\prod_{t=1}^{T}p_\theta (x_{t-1}^{adv}|x_{t}^{adv},x)
		\end{aligned}
		\label{eq:cdma_sample}
	\end{equation}
	Here each transition $p_\theta(x_{t-1}^{adv}|x_{t}^{adv},x)$ in the sampling process depends on the condition $x$, i.e., the clean image. The sampling (Eq.\ref{eq:ddpm_sample_1}) in the conditional version is re-written as:
	
	\begin{equation}
		p_\theta (x_{t-1}^{adv}|x_{t}^{adv},x)=\mathcal{N} (x_{t-1}^{adv};\mu _\theta (x_{t}^{adv},t,x) , {\textstyle \sum_{\theta }^{}(x_{t}^{adv},t,x))}
		\label{alg:c_reverse}
	\end{equation}
	As shown in Eq. \ref{alg:c_reverse}, \Name generates the adversarial example $\bm{x}^{adv}$ via the diffusion model's reverse Markov process and starts from $\bm{x}_{0}^{adv} =\epsilon \sim \mathcal{N}(0,\mathrm{I})$ with the conditional clean image $x$. To make the final adversarial examples meet the similarity requirements, we impose the extra $clip(\cdot)$ constraints on $L_{\infty}$-norm as:
	\begin{equation}
		\bm{x}_{final}^{adv} = clip(clip(\bm{x}_{0}^{adv},x-\epsilon,x+\epsilon),0,1)
	\end{equation}
	where $\epsilon $ is the adversarial perturbation budget.
	
	The training and attacking algorithms of \Name are listed in Alg. \ref{alg:alg1} and Alg. \ref{alg:alg2}, respectively, which could help readers to re-implement our method step-by-step.

	
	\begin{table*}[t]    
		\caption{Experimental results on attack success rate and the query counts on CIFAR-10.}
		\label{tab:cifar-10}
		\centering
		\small
		\renewcommand{\arraystretch}{1.1}
		\setlength\tabcolsep{5.5pt}

		\begin{tabular}{c|c|ccc|ccc|ccc|ccc}
			\hline        
			\multirow{2}{*}{}                           & \multirow{2}{*}{Methods} & \multicolumn{3}{c|}{VGG-19} & \multicolumn{3}{c|}{Inception-V3} & \multicolumn{3}{c|}{ResNet-50} & \multicolumn{3}{c}{DenseNet-169}                                                                                                                          \\
			&                          & ASR                         & Avg.Q                             & Med.Q                          & ASR                              & Avg.Q         & Med.Q      & ASR            & Avg.Q         & Med.Q      & ASR            & Avg.Q         & Med.Q      \\
			\hline
			\multirow{10}{*}{\rotatebox{90}{untarget}} & AdvFlow                  & 80.60                       & 396.79                            & 358                            & 61.5                             & 423.68        & 358        & 63.20          & 403.76        & 358        & 65.87          & 440.29        & 409        \\
			& RayS                     & 98.93                       & 160.25                            & 126                            & 96.01                            & 224.52        & 176.5      & 96.53          & 202.43        & 150        & 96.65          & 215.62        & 159        \\
			& Bayes\_Attack            & 75.24                       & 45.87                             & 5                              & 79.39                            & 48.2          & 5          & 81.44          & 42.37         & 5          & 77.10          & 45.42         & 5          \\
			& TA                       & 25.05                       & 66.60                             & 11                             & 23.77                            & 67.01         & 5          & 28.60          & 74.02         & 6          & 27.69          & 66.29         & 8          \\
			& NPAttack                 & 97.75                       & 225.79                            & 150                            & 96.73                            & 229.3         & 150        & 96.11          & 234.72        & 150        & 96.54          & 239.45        & 150        \\
			& ODS                      & 97.49                       & 11.79                             & 10                             & 99.00                            & 28.50         & 13         & 97.80          & 18.03         & 10         & 99.1           & 25.26         & 12         \\
			& GFCS                     & 98.83                       & 8.66                              & 6                              & 98.29                            & 28.02         & 7          & 98.43          & 10.38         & 6          & 99.45          & 18.05         & 7          \\
			& CG-Attack                & 96.57                       & 86.94                             & \textbf{1}                     & 97.98                            & 101.65        & \textbf{1} & 96.48          & 84.97         & \textbf{1} & 97.16          & 97.91         & \textbf{1} \\
			& MCG-Attack               & 96.87                       & 75.41                             & \textbf{1}                     & 98.14                            & 80.67         & \textbf{1} & 97.38          & 76.92         & \textbf{1} & 99.10          & 81.67         & \textbf{1} \\
			& CDMA(Ours)               & \textbf{99.46}              & \textbf{1.63}                     & \textbf{1}                     & \textbf{99.58}                   & \textbf{2.97} & \textbf{1} & \textbf{99.58} & \textbf{3.16} & \textbf{1} & \textbf{99.68} & \textbf{5.31} & \textbf{1} \\
			\hline
			\multirow{10}{*}{\rotatebox{90}{targeted}}  & AdvFlow                  & 10.93                       & 653.12                            & 600                            & 9.16                             & 674.17        & 650        & 10.64          & 672.13        & 650        & 9.87           & 681.67        & 650        \\
			& RayS                     & 18.82                       & 209.95                            & 159                            & 18.40                            & 331.5         & 283        & 17.82          & 297.91        & 258        & 17.25          & 306.82        & 222        \\
			& Bayes\_Attack            & 28.62                       & 587.16                            & 490                            & 24.94                            & 505.61        & 535        & 19.87          & 549.48        & 570        & 27.84          & 497.99        & 515        \\
			& TA                       & 14.63                       & 238.65                            & 210                            & 13.58                            & 267.21        & 231        & 15.87          & 227.41        & 204        & 14.92          & 219.36        & 198        \\
			& NPAttack                 & 48.71                       & 279.16                            & 250                            & 51.87                            & 278.13        & 200        & 47.40          & 400           & 450        & 42.88          & 322.92        & 300        \\
			& ODS                      & 76.71                       & 204.08                            & 80                             & 88.42                            & 152.17        & 75         & 92.01          & 130.81        & 66         & 95.08          & 133.14        & 70         \\
			& GFCS                     & 79.08                       & 141.88                            & 23                             & 90.68                            & 111.9         & 32         & 93.39          & 95.53         & 24.5       & 95.40          & 99.41         & 25         \\
			& CG-Attack                & 74.03                       & 487.61                            & 441                            & 78.67                            & 511.13        & 501        & 76.94          & 491.67        & 481        & 77.15          & 534.16        & 501        \\
			& MCG-Attack               & 79.17                       & 361.47                            & 281                            & 81.92                            & 306.28        & 261        & 80.69          & 297.63        & 261        & 78.16          & 342.19        & 301        \\
			& CDMA(Ours)               & \textbf{94.31}              & \textbf{27.85}                    & \textbf{1}                     & \textbf{91.67}                   & \textbf{8.88} & \textbf{1} & \textbf{94.65} & \textbf{7.47} & \textbf{1} & \textbf{95.52} & \textbf{8.09} & \textbf{1} \\
			\hline
		\end{tabular}
		
	\end{table*}

	
	\begin{table*}[!ht]
		\caption{Experimental results on ASR and the query counts on CIFAR-100.}
		\label{tab:cifar-100}
		\centering
		\small
		\renewcommand{\arraystretch}{1.1}
		\setlength\tabcolsep{5.5pt}
		\begin{tabular}{c|c|ccc|ccc|ccc|ccc}
			\hline
			\multirow{2}{*}{}                           & \multirow{2}{*}{Methods} & \multicolumn{3}{c|}{VGG-19} & \multicolumn{3}{c|}{Inception-V3} & \multicolumn{3}{c|}{ResNet-50} & \multicolumn{3}{c}{DenseNet-169}                                                                                                                             \\
			&                          & ASR                         & Avg.Q                             & Med.Q                          & ASR                              & Avg.Q          & Med.Q      & ASR            & Avg.Q          & Med.Q      & ASR            & Avg.Q          & Med.Q      \\
			\hline
			\multirow{10}{*}{\rotatebox{90}{untarget}} & AdvFlow                  & 75.30                       & 321.02                            & 256                            & 79.00                            & 392.24         & 358        & 76.10          & 365.17         & 307        & 81.80          & 354.82         & 307        \\
			& RayS                     & \textbf{99.86}              & 101.28                            & 71                             & 98.63                            & 134.8          & 96         & 99.47          & 130.87         & 82         & 98.76          & 130.39         & 84.5       \\
			& Bayes\_Attack            & 89.70                       & 13.82                             & 5                              & 89.03                            & 14.8           & 5          & 88.76          & 17.24          & 5          & 87.58          & 18.69          & 5          \\
			& TA                       & 55.96                       & 38.85                             & 5                              & 48.82                            & 38.57          & 5          & 53.39          & 43.54          & 5          & 54.42          & 57.27          & 6          \\
			& NPAttack                 & 94.74                       & 161.34                            & 100                            & 94.11                            & 172.84         & 100        & 95.39          & 173.39         & 100        & 94.61          & 174.54         & 100        \\
			& ODS                      & 97.90                       & 35.84                             & 17                             & 97.59                            & 28.89          & 19         & 96.78          & 32.60          & 18         & 96.72          & 28.93          & 17         \\
			& GFCS                     & 98.84                       & 8.66                              & 6                              & 98.27                            & 28.02          & 7          & 97.97          & 10.38          & 6          & 97.49          & 18.05          & 7          \\
			& CG--Attack               & 98.74                       & 78.94                             & \textbf{1}                     & 87.59                            & 97.38          & 21         & 98.27          & 79.14          & \textbf{1} & 97.76          & 96.83          & \textbf{1} \\
			& MCG-Attack               & 98.87                       & 69.19                             & \textbf{1}                     & 90.64                            & 84.61          & \textbf{1} & 98.82          & 71.66          & \textbf{1} & 98.47          & 85.73          & \textbf{1} \\
			& CDMA(Ours)               & 99.25                       & \textbf{5.16}                     & \textbf{1}                     & \textbf{98.71}                   & \textbf{6.86}  & \textbf{1} & \textbf{99.63} & \textbf{4.28}  & \textbf{1} & \textbf{99.37} & \textbf{4.92}  & \textbf{1} \\
			\hline
			\multirow{10}{*}{\rotatebox{90}{targeted}}  & AdvFlow                  & 7.12                        & 697.58                            & 650                            & 8.32                             & 681.65         & 650        & 7.98           & 657.29         & 600        & 8.42           & 642.13         & 600        \\
			& RayS                     & 13.44                       & 269.90                            & 182                            & 12.74                            & 232.17         & 267        & 14.05          & 216.125        & 194        & 11.12          & 212.67         & 181        \\
			& Bayes\_Attack            & 16.94                       & 645.61                            & 710                            & 13.48                            & 687.15         & 695        & 15.49          & 597.34         & 625        & 17.61          & 578.67         & 600        \\
			& TA                       & 13.67                       & 281.52                            & 247                            & 11.92                            & 277.40         & 239        & 12.00          & 274.58         & 235        & 13.94          & 264.59         & 232        \\
			& NPAttack                 & 40.59                       & 451.63                            & 400                            & 41.25                            & 318.52         & 300        & 42.13          & 487.35         & 450        & 40.38          & 507.49         & 550        \\
			& ODS                      & 75.78                       & 301.89                            & 224.5                          & 76.95                            & 224.11         & 157        & 79.51          & 291.93         & 208        & 73.67          & 257.95         & 187        \\
			& GFCS                     & 74.53                       & 292.10                            & 198                            & 80.75                            & 155.17         & 68.5       & 81.66          & 181.79         & 75         & 78.30          & 164.37         & 71         \\
			& CG-Attack                & 62.91                       & 648.29                            & 601                            & 60.74                            & 676.49         & 621        & 63.15          & 542.67         & 481        & 61.94          & 704.61         & 641        \\
			& MCG-Attack               & 66.28                       & 546.58                            & 481                            & 62.18                            & 516.94         & 441        & 67.16          & 486.27         & 441        & 63.35          & 536.93         & 501        \\
			& CDMA(Ours)               & \textbf{73.95}              & \textbf{70.17}                    & \textbf{2}                     & \textbf{83.42}                   & \textbf{40.43} & \textbf{1} & \textbf{82.79} & \textbf{24.29} & \textbf{1} & \textbf{77.25} & \textbf{38.10} & \textbf{1} \\
			\hline
		\end{tabular}
		
	\end{table*}
	
	
	\begin{table*}[!ht]
		\caption{Experimental results on attack success rate and the query counts on Tiny-ImageNet.}
		\label{tab:tiny-imagenet-200}
		\centering
		\small
		\renewcommand{\arraystretch}{1.1}
		\setlength\tabcolsep{5.5pt}
		\begin{tabular}{c|c|ccc|ccc|ccc|ccc}
			\hline
			\multirow{2}{*}{}                           & \multirow{2}{*}{Methods} & \multicolumn{3}{c|}{VGG-19} & \multicolumn{3}{c|}{Inception-V3} & \multicolumn{3}{c|}{ResNet-50} & \multicolumn{3}{c}{DenseNet-169}                                                                                                                             \\
			&                          & ASR                         & Avg.Q                             & Med.Q                          & ASR                              & Avg.Q          & Med.Q      & ASR            & Avg.Q          & Med.Q      & ASR            & Avg.Q          & Med.Q      \\
			\hline
			\multirow{10}{*}{\rotatebox{90}{untarget}} & AdvFlow                  & 88.30                       & 302.44                            & 256                            & 92.89                            & 322.28         & 256        & 93.50          & 313.44         & 256        & 98.8           & 241.86         & 205        \\
			& RayS                     & 98.98                       & 100.2                             & 68                             & 99.39                            & 117.83         & 68         & 98.77          & 107.73         & 67         & 99.64          & 93.5           & 65         \\
			& Bayes\_Attack            & 74.13                       & 24.14                             & 5                              & 76.28                            & 66.13          & 5          & 72.41          & 24.83          & 5          & 80.00          & 26.37          & 5          \\
			& TA                       & 69.32                       & 73.62                             & 18                             & 64.72                            & 82.41          & 18         & 67.76          & 71.74          & 16         & 78.29          & 59.53          & 11         \\
			& NPAttack                 & 91.77                       & 241.55                            & 150                            & 93.77                            & 259.48         & 150        & 95.59          & 242.68         & 150        & 98.83          & 192.16         & 150        \\
			& ODS                      & 99.61                       & 43.47                             & 24                             & 99.17                            & 45             & 30         & 99.53          & 42.28          & 27         & 98.82          & 47.43          & 25.5       \\
			& GFCS                     & 98.41                       & 36.64                             & 9                              & 99.37                            & 35             & 11         & 99.21          & 36.38          & 10         & 99.67          & 25.12          & 8          \\
			& CG-Attack                & 98.34                       & 97.81                             & 21                             & 97.73                            & 113.94         & 21         & 97.62          & 136.81         & 21         & 98.16          & 127.43         & 21         \\
			& MCG-Attack               & 98.76                       & 80.64                             & 1                              & 98.47                            & 110.49         & 21         & 97.91          & 109.84         & 21         & 99.17          & 89.76          & 21         \\
			& CDMA(Ours)               & \textbf{99.67}              & \textbf{3.90}                     & \textbf{1}                     & \textbf{99.55}                   & \textbf{7.27}  & \textbf{1} & \textbf{99.71} & \textbf{5.53}  & \textbf{1} & \textbf{99.83} & \textbf{3.75}  & \textbf{1} \\
			\hline
			\multirow{10}{*}{\rotatebox{90}{targeted}}  & AdvFlow                  & 5.32                        & 754.38                            & 700                            & 5.73                             & 724.69         & 700        & 6.52           & 671.94         & 650        & 5.84           & 714.61         & 700        \\
			& RayS                     & 9.27                        & 255.57                            & 215                            & 8.89                             & 223.17         & 197        & 10.45          & 192.33         & 223        & 9.72           & 217.70         & 184.5      \\
			& Bayes\_Attack            & 10.94                       & 284.64                            & 245                            & 13.14                            & 297.92         & 305        & 12.49          & 273.25         & 280        & 11.76          & 318.16         & 315        \\
			& TA                       & 8.64                        & 263.59                            & 243.5                          & 7.48                             & 316.78         & 284        & 8.12           & 323.43         & 251        & 8.47           & 297.69         & 267        \\
			& NPAttack                 & 32.26                       & 585.71                            & 600                            & 32.12                            & 605.35         & 625        & 36.69          & 596.02         & 650        & 33.90          & 690.52         & 600        \\
			& ODS                      & 74.09                       & 391.22                            & 271                            & 81.01                            & 336.33         & 286        & 81.80          & 340.50         & 280.5      & 80.64          & 341.62         & 252.5      \\
			& GFCS                     & \textbf{79.37}              & 298.25                            & 235                            & 81.54                            & 220.46         & 110.5      & 84.83          & 250.64         & 142        & 82.34          & 326.30         & 157        \\
			& CG-Attack                & 71.67                       & 367.45                            & 381                            & 69.46                            & 416.97         & 401        & 73.97          & 437.41         & 421        & 68.63          & 443.37         & 421        \\
			& MCG-Attack               & 70.16                       & 417.28                            & 401                            & 71.91                            & 429.71         & 401        & 75.63          & 431.67         & 421        & 70.38          & 427.16         & 401        \\
			& CDMA(Ours)               & 77.95                       & \textbf{46.04}                    & \textbf{1}                     & \textbf{82.20}                   & \textbf{26.87} & \textbf{1} & \textbf{85.13} & \textbf{17.80} & \textbf{1} & \textbf{82.82} & \textbf{23.40} & \textbf{1} \\
			\hline
		\end{tabular}
		
	\end{table*}

	\section{Evaluation}
	\label{Sec:experiments}
	We present the experimental results of \Name. We first compare it with other black-box attack baselines in untarget and targeted scenarios. Then we measure the attack effectiveness against state-of-the-art defenses. Next, we show the results of data-independent and model-independent attacks. Finally, we show the ablation study results to explore the attack ability of \Name under different settings.
	
	\subsection{Experimental Setup}
	\noindent\textbf{Implementation.} We set the maximum number of queries as $Q_{max}=1000$ to simulate a realistic attack scenario. We stop the attack once a specific input is mispredicted by the victim model successfully. We set the noise budget as $ \epsilon =8/255.$ and $ \epsilon =16/255.$, which is shortened as $ \epsilon =8$ and $ \epsilon =16$  for all attacks. To train the diffusion model in \Name, the total number of diffusion steps is $T=2000$. The number of training epochs is $E=1e8$ with the batch size of $B=256$. The noise scheduler is "linear" starting from $1e-6$ and ending with $0.01$. All the experiments are conducted on a GPU server with 4*NVIDIA Tesla A100 40GB GPU, 2*Xeon Glod 6112 CPU and RAM 512GB.

	\noindent\textbf{Datasets.} We verify the performance of \Name on three benchmark datasets for computer vision task, named CIFAR-10\footnote{http://www.cs.toronto.edu/~kriz/cifar.html} \cite{cifar-10}, CIFAR-100\footnote{http://www.cs.toronto.edu/~kriz/cifar.html} \cite{cifar-10} and Tiny-ImageNet-200\footnote{http://cs231n.stanford.edu/tiny-imagenet-200.zip} \cite{cvpr/DengDSLL009}. In detail, CIFAR-10 contains 50,000 training images and 10,000 testing images with the size of $3\times32\times32$ from 10 classes; CIFAR-100 has 100 classes, including the same number of training and testing images as the CIFAR-10; Tiny-ImageNet-200 has 200 categories, containing about 1.3M samples for training and 50,000 samples for validation. In our experiments, we first generate adversarial examples using white-box attacks on the training set of the above three datasets to train the diffusion model, and then randomly sample 1,000 images from the test set of these datasets for attacks.

	\noindent\textbf{Models.} We train a few widely-used deep neural networks, including VGG \cite{corr/SimonyanZ14a}, Inception \cite{cvpr/SzegedyVISW16,aaai/SzegedyIVA17}, ResNet \cite{cvpr/HeZRS16}, and DenseNets \cite{cvpr/HuangLMW17} over the aforementioned datasets until the models achieve the best classification results. Among them, We adopt VGG-13, ResNet-18 and DeseNet-121 as the shadow models for all query- and transfer-based baselines and \Name, while VGG-19, Inception-V3, ResNet-50 and DenseNet-169 as the victim models to be attacked for all the methods. The  top-1 classification accuracy of these victim models are 90.48\%, 84.51\%, 94.07\%, and 94.24\% for CIFAR-10, 66.81\%, 77.86\%, 76.05\% and 77.18\% for CIFAR-100 and 57.62\%, 65.89\%, 65.41\% and 56.04\% for Tiny-ImageNet-200, respectively.
	
	\noindent\textbf{Baselines.}
	We select nine state-of-the-art black-box attacks as the baselines, including score-based, decision-based, and query- and transfer-based methods. These include Rays \cite{kdd/ChenG20}, AdvFlow \cite{nips/DolatabadiEL20}, Bayes\_Attack \cite{kdd/ShuklaSWK21}, TA \cite{eccv/WangZTGHLL22}, NPAttack \cite{pr/BaiWZJX23}, ODS \cite{nips/Tashiro0E20}, GFCS \cite{iclr/LordMB22}, CG-Attack \cite{cvpr/FengWFL0X22} and MCG-Attack \cite{corr/abs-2301-00364}. We reproduce the attacks from the code released in the original papers with the default settings.
	
	\noindent\textbf{Metrics.} We perform evaluations with the following metrics: Attack Success Rate (ASR) measures the attack effectiveness. Average and Median numbers of queries (Avg.Q and Med.Q) measure the attack efficiency.
	
	\subsection{Comparisons with Baseline Attacks}
	\label{subsec:attack}
	Tables \ref{tab:cifar-10}, \ref{tab:cifar-100} and \ref{tab:tiny-imagenet-200} present the untarget and targeted attack performance comparison with all baselines under the noise budget $\epsilon=16$ on VGG-19, Inception-V3, ResNet-50, and DenseNet-169, respectively. Specifically, in both untarget and targeted situations, we observe that our proposed \Name enjoys much higher efficiency in terms of the average and median numbers of queries, as well as much higher attack success rate than AdvFlow, Bayes\_Attack and TA for all datasets. Compared to the rest baselines, although the attack success rate of \Name does not exceed too much, in some cases, even lower than Rays (when extending the untarget attack on VGG-19 with CIFAR-100) and GFCS (when extending the targeted attack on VGG-19 with Tiny-ImageNet-200), the Avg.Q and Med.Q are always lowest than all methods, especially in the target setting. \Name only needs several queries to obtain a near 100\% attack success rate and the Med.Q of \Name is 1. These experimental results demonstrate the superiority of our proposed method in terms of attack effectiveness and efficiency. 
	

	Table. \ref{tab:un_eps_8} presents the performance comparison of all attack baselines on VGG-19, Inception-V3, ResNet-50 and DenseNet-169, respectively, where the noise budget is set to $\epsilon=8$. Although the attack becomes more challenging with a small noise budget, compared with all the attack baselines, the proposed \Name can also get the best attack performance in most situations. Especially the average and median queries of \Name are still the lowest in all cases, which have exhibited the high effectiveness of the proposed methods.

	
	\begin{table*}
		\caption{untarget attack on CIFAR-10, CIFAR-100 and Tiny-ImageNet, the noise budget is $\epsilon=8$.}
		\label{tab:un_eps_8}
		\centering
		\small
		\renewcommand{\arraystretch}{1.1}
		\setlength\tabcolsep{5.5pt}
		\begin{tabular}{c|c|ccc|ccc|ccc|ccc}
			\hline
			\multirow{2}{*}{}  & \multirow{2}{*}{Methods} & \multicolumn{3}{c|}{VGG-19} & \multicolumn{3}{c|}{Inception-V3} & \multicolumn{3}{c|}{ResNet-50} & \multicolumn{3}{c}{DenseNet-169}                                                                                                                             \\
			&                          & ASR                         & Avg.Q                             & Med.Q                          & ASR                              & Avg.Q          & Med.Q      & ASR            & Avg.Q          & Med.Q      & ASR            & Avg.Q          & Med.Q      \\
			\hline
			\multirow{10}{*}{\rotatebox{90}{CIFAR-10}} & AdvFlow                  & 50.1                        & 414.19                            & 358                            & 37.43                            & 431.9          & 409        & 41.50          & 409.62         & 358        & 39.72          & 438.49         & 409        \\
			& RayS                     & 80.96                       & 303.45                            & 231                            & 75.08                            & 342.30         & 286        & 79.06          & 339.32         & 254        & 77.14          & 331.54         & 255        \\
			& Bayes\_Attack            & 33.73                       & 91.57                             & 8                              & 36.80                            & 95.01          & 5          & 38.92          & 84.93          & 7          & 34.98          & 78.45          & 6          \\
			& TA                       & 7.56                        & 95.39                             & 14                             & 5.60                             & 63.24          & 5          & 6.37           & 52.58          & 7          & 7.16           & 87.95          & 9.5        \\
			& NPAttack                 & 64.67                       & 330.30                            & 250                            & 64.17                            & 336.37         & 250        & 65.90          & 328.98         & 250        & 65.19          & 339.95         & 250        \\
			& ODS                      & \textbf{95.16}              & 11.71                             & 9                              & 92.10                            & 25.19          & 12         & 96.24          & 15.84          & 10         & 95.80          & 26.28          & 12         \\
			& GFCS                     & 94.7                        & 25.49                             & 6                              & 91.30                            & 37.83          & 7          & 96.20          & 22.27          & 6          & 94.80          & 29.03          & 7          \\
			& CG-Attack                & 94.84                       & 125.95                            & \textbf{1}                     & 92.94                            & 137.64         & 41         & 94.87          & 115.57         & \textbf{1} & 94.91          & 134.63         & \textbf{1} \\
			& MCG-Attack               & 95.07                       & 105.63                            & \textbf{1}                     & 93.25                            & 110.91         & \textbf{1} & 95.71          & 99.14          & \textbf{1} & 95.68          & 100.84         & \textbf{1} \\
			& CDMA(Ours)               & 94.36                       & \textbf{9.38}                     & \textbf{1}                     & \textbf{94.08}                   & \textbf{16.66} & \textbf{1} & \textbf{96.66} & \textbf{10.83} & \textbf{1} & \textbf{96.50} & \textbf{10.78} & \textbf{1} \\
			\hline
			\multirow{10}{*}{\rotatebox{90}{CIFAR-100}} & AdvFlow                  & 58.68                       & 340.3                             & 307                            & 53.10                            & 375.77         & 307        & 52.41          & 383.79         & 358        & 55.30          & 369.44         & 307        \\
			& RayS                     & \textbf{94.54}              & 197.51                            & 135.5                          & 86.93                            & 227.89         & 158        & 92.00          & 214.4          & 139.5      & 91.19          & 222.94         & 148.5      \\
			& Bayes\_Attack            & 65.16                       & 36.62                             & 5                              & 65.21                            & 46.50          & 5          & 63.86          & 45.34          & 5          & 62.61          & 56.06          & 5          \\
			& TA                       & 22.43                       & 63.26                             & 5                              & 17.45                            & 58.55          & 5          & 22.60          & 66.49          & 5          & 21.35          & 64.26          & 5          \\
			& NPAttack                 & 71.20                       & 236.55                            & 150                            & 70.43                            & 253.03         & 150        & 70.13          & 240.68         & 150        & 69.13          & 271.60         & 150        \\
			& ODS                      & 93.90                       & 33.13                             & 11                             & 94.65                            & 18.89          & 12         & 92.70          & 22.00          & 12         & 91.70          & 28.27          & 14         \\
			& GFCS                     & 94.20                       & 33.44                             & 5                              & \textbf{95.50}                   & 30.65          & 6          & 93.60          & 30.42          & 6          & 92.00          & 29.41          & 6          \\
			& CG-Attack                & 93.67                       & 112.67                            & 21                             & 93.47                            & 138.59         & 21         & 93.29          & 100.93         & 41         & 92.61          & 125.48         & 41         \\
			& MCG-Attack               & 94.15                       & 105.64                            & \textbf{1}                     & 94.39                            & 110.94         & \textbf{1} & 93.68          & 89.14          & \textbf{1} & 93.62          & 98.07          & \textbf{1} \\
			& CDMA(Ours)               & 92.42                       & \textbf{16.17}                    & \textbf{1}                     & 90.70                            & \textbf{19.28} & \textbf{1} & \textbf{94.75} & \textbf{18.17} & \textbf{1} & \textbf{92.07} & \textbf{20.20} & \textbf{1} \\
			\hline
			\multirow{10}{*}{\rotatebox{90}{Tiny-ImageNet}} & AdvFlow                  & 76.98                       & 319.2                             & 256                            & 79.96                            & 334.42         & 256        & 82.3           & 319.08         & 256        & 95.70          & 273.27         & 205        \\
			& RayS                     & 89.51                       & 185.96                            & 124.5                          & 87.97                            & 199.93         & 132        & 88.10          & 170.94         & 109        & 93.56          & 160.74         & 98         \\
			& Bayes\_Attack            & 43.58                       & 59.27                             & 5                              & 40.18                            & 77.91          & 5          & 42.23          & 82.09          & 5          & 55.36          & 53.51          & 5          \\
			& TA                       & 34.66                       & 100.99                            & 26                             & 29.29                            & 107.02         & 30.5       & 33.07          & 103.05         & 24.5       & 49.61          & 98.07          & 29         \\
			& NPAttack                 & 71.57                       & 315.75                            & 250                            & 73.32                            & 315.14         & 250        & 74.46          & 317.67         & 250        & 88.70          & 57.39          & 150        \\
			& ODS                      & 91.50                       & 42.55                             & 23                             & 91.21                            & 45.47          & 28         & 91.17          & 44.10          & 28         & 93.31          & 42.33          & 28         \\
			& GFCS                     & 92.70                       & 62.18                             & 9                              & 93.40                            & 71.52          & 12         & 92.10          & 79.95          & 12         & 95.00          & 49.39          & 9          \\
			& CG-Attack                & 92.64                       & 111.42                            & 21                             & 87.65                            & 164.97         & 41         & 88.49          & 157.49         & 41         & 90.48          & 146.73         & 21         \\
			& MCG-Attack               & 92.93                       & 104.91                            & 21                             & 89.46                            & 146.18         & 41         & 89.07          & 139.73         & 21         & 91.86          & 124.61         & 21         \\
			& CDMA(Ours)               & \textbf{93.57}              & \textbf{19.69}                    & \textbf{1}                     & \textbf{93.53}                   & \textbf{31.98} & \textbf{1} & \textbf{92.13} & \textbf{27.26} & \textbf{1} & \textbf{95.34} & \textbf{11.63} & \textbf{1} \\
			\hline
		\end{tabular}
		
	\end{table*}

\begin{figure*}[t]
		\centering
		\includegraphics[width=0.98\textwidth]{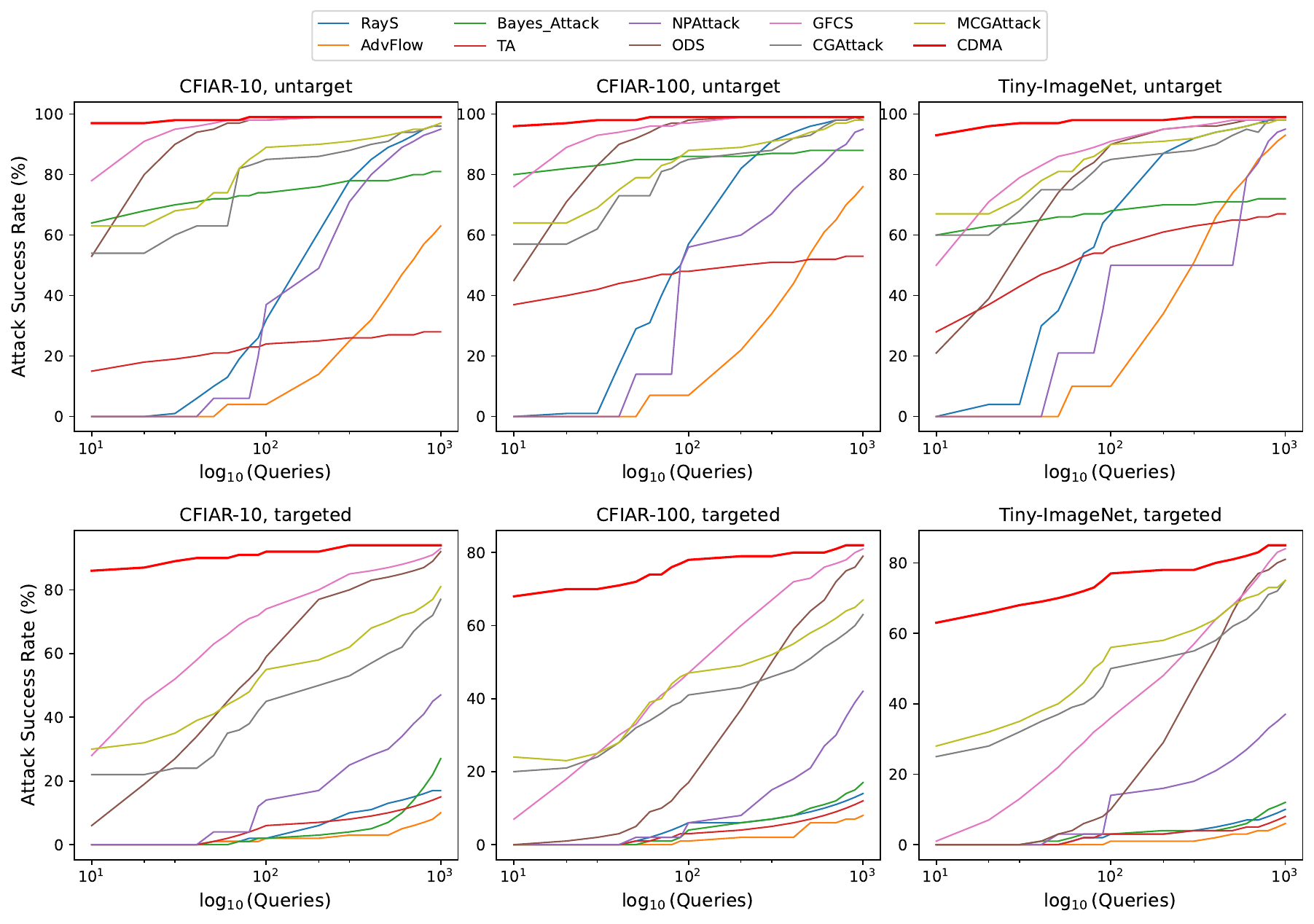}
		\caption{Queries vs. ASR on CIFAR-100 and Tiny-ImageNet for untarget and targeted attack settings. The maximal query counts are limited to 1000 and the noise budget's $L_{\infty} $ norm is set to $ \epsilon = 16$.}
		\label{fig:asr_query}
	\end{figure*}

	Figure~\ref{fig:asr_query} shows the attack success rate versus the number of queries for all baseline methods over CIFAR-10, CIFAR-100, and Tiny-ImageNet-200 in untarget and target attack settings. Again we can see that \Name achieves the highest attack success rate in most situations and the best query efficiency compared with other black-box attack baselines. The results show that our proposed \Name can achieve the highest attack success rate under the same query counts. Note that \Name can obtain a boosting attack success rate at the first few queries, especially under targeted attack settings, while other attacks only can obtain a satisfactory attack success rate after hundreds of queries.
	
	\subsection{Adversarial Robustness to Defense Strategies}
	To further evaluate the generated adversarial examples' robustness, we adopt some defense methods to purify or pre-process the malicious examples, and then measure their effectiveness. The defense methods in our consideration involve JPEG compression (JPEG) \cite{shin2017jpeg}, NRP \cite{nips/Naseer0KKP19}, pixel deflection (PD) \cite{cvpr/PrakashMGDS18}, GuidedDiffusionPur (GDP) \cite{corr/abs-2205-14969}, RP-Regularizer (RP) \cite{corr/abs-1810-00953}, BitDepthReduce (BDR) \cite{ndss/Xu0Q18} and MedianSmoothing2D (MS) \cite{ndss/Xu0Q18}. We first synthesize adversarial examples on ResNet-50 for CIFAR-10, and then measure their attack success rate against these defense strategies. The results are shown in Table. \ref{fig:resistence}. Among all the black-box attack methods, our proposed method has the highest attack success rate in most cases, which implies the adversarial examples generated by \Name are more robust to current defense methods compared with other attacks.

	\subsection{Data-independent and Model-independent Attacks}
	\subsubsection{\textbf{Data-independent Attack}} We carry out attacks across different datasets to verify the generalization of our \Name. The datasets include STL-10 \cite{jmlr/CoatesNL11}, Caltech-256 \cite{cviu/Fei-FeiFP07}, Places-365 \cite{zhou2017places}) and CeleBA \cite{iccv/LiuLWT15}. These datasets are not used for training the diffusion model, and we only sample images from their test set to test the effectiveness of \Name.
	
	\begin{table}[t]
		\centering       
		\caption{The attack success rate under defense strategies.}
		\label{fig:resistence}
		\footnotesize
		\renewcommand{\arraystretch}{1.1}
		\setlength\tabcolsep{4pt}
		\begin{tabular}{cccccccc}
			\hline
			Methods      & JPEG           & NRP            & PD             & GDP            & RP             & BDR            & MS             \\
			\hline
			AdvFlow       & 46.36          & 44.46          & 87.15          & 27.30          & 61.86          & 46.79          & 52.88          \\
			RayS          & 60.99          & 15.73          & 90.07          & 21.65          & 63.36          & 57.44          & 58.73          \\
			Bayes\_Attack & 43.36          & 22.01          & 87.89          & 25.52          & 66.02          & 80.99          & 49.35          \\
			TA            & 42.91          & 33.20          & 82.81          & 21.88          & 53.82          & 71.88          & 45.49          \\
			NPAttack      & 38.93          & 31.84          & 90.40          & 11.61          & 64.25          & 59.54          & 41.89          \\
			ODS           & 21.08          & 41.74          & \textbf{91.10} & 4.91           & 66.42          & 41.84          & 30.19          \\
			GFCS          & 21.72          & 42.34          & 89.79          & 6.36           & 66.73          & 44.35          & 29.23          \\
			CG-Attack     & 47.61          & 38.72          & 78.16          & 18.67          & 59.86          & 68.29          & 52.63          \\
			M-CG-Attack   & 48.43          & 40.81          & 82.69          & 22.38          & 60.53          & 69.62          & 54.27          \\
			DMA(Ours)     & \textbf{72.61} & \textbf{77.01} & 90.40          & \textbf{44.20} & \textbf{93.86} & \textbf{91.53} & \textbf{80.19} \\
			\hline
		\end{tabular}    
	\end{table}
	
	\begin{table}[ht]
		\caption{The attack success rate of data-independent attack.}
		\label{tab:cross}
		\centering
		\small
		\renewcommand{\arraystretch}{1.1}
		\setlength\tabcolsep{3pt}
		\resizebox{1\linewidth}{!}{
			
			\begin{tabular}{c|c|cc|cc|cc|ll} 
				\hline
				\multirow{2}{*}{Noise}         & \multirow{2}{*}{Models} & \multicolumn{2}{c|}{STL-10} & \multicolumn{2}{c|}{Places-365} & \multicolumn{2}{c|}{Caltech256} & \multicolumn{2}{l}{CeleBA}  \\
				&                         & ASR   & Avg.Q               & ASR  & Avg.Q                    & ASR  & Avg.Q                    & ASR  & Avg.Q                \\ 
				\hline
				\multirow{2}{*}{$\epsilon=8$}  & VGG-19                  & 94.29 & 10.15               & 99.5 & 4.47                     & 96.9 & 11.94                    & 98.2 & 9.43                 \\
				& ResNet-50               & 95.95 & 15.13               & 99.6 & 4.55                     & 97.9 & 9.58                     & 99.7 & 13.57                \\ 
				\hline
				\multirow{2}{*}{$\epsilon=16$} & VGG-19                  & 99.41 & 4.38                & 99.9 & 2.02                     & 99.7 & 1.99                     & 100  & 2.18                 \\
				& ResNet-50               & 100   & 6.43                & 100  & 1.56                     & 99.9 & 3.00                     & 100  & 2.21                 \\
				\hline
			\end{tabular}
		}
	\end{table}
	
\begin{figure}[t]
		\centering
		
		\includegraphics[width=0.48\textwidth]{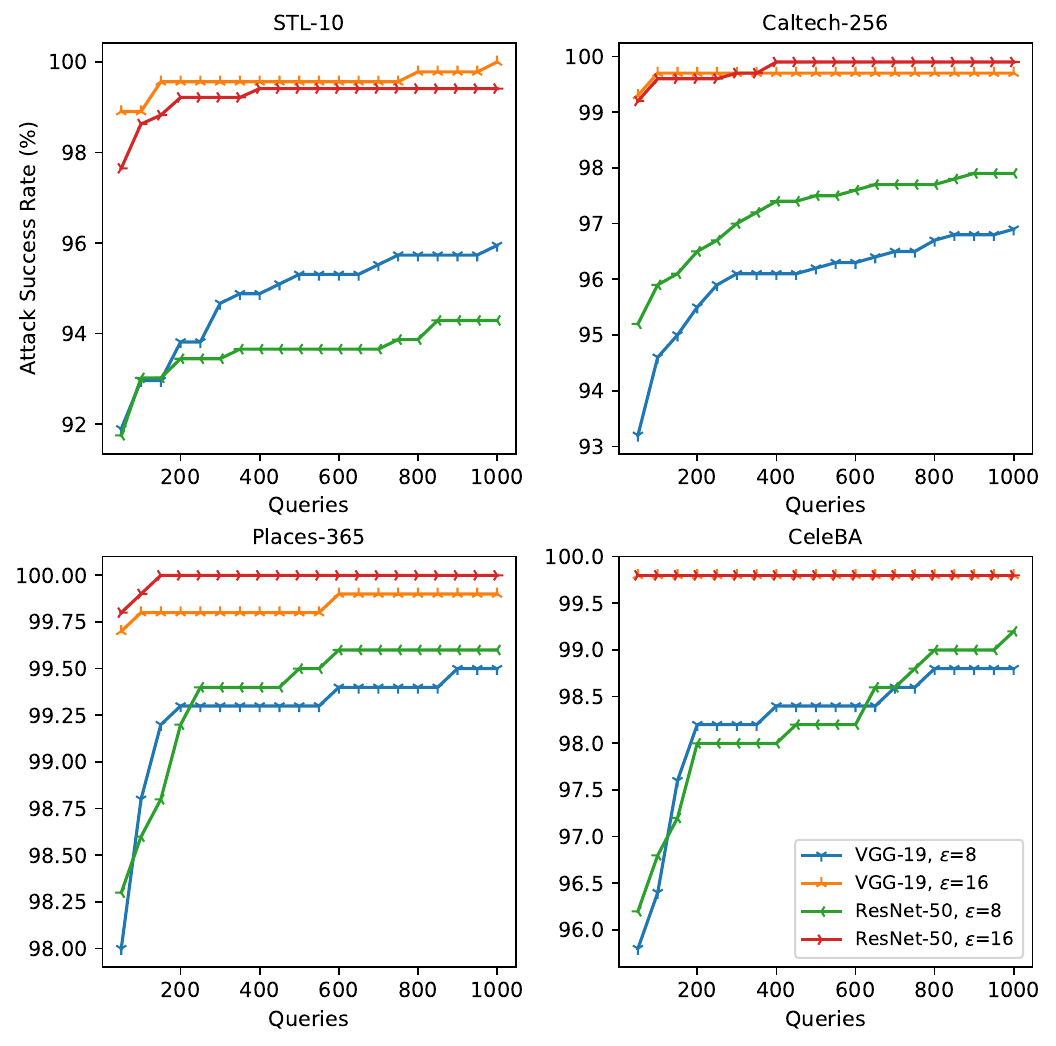}
		\setlength{\abovecaptionskip}{3pt}
		\setlength{\belowcaptionskip}{-15pt}
		\caption{ASR vs. Queries under data-independent settings.}
		
		\label{fig:cross}
	\end{figure}

	In detail, we train a diffusion model based on a specific dataset (CFAIR-10 and Tiny-ImageNet-200) and then apply the attack on another dataset (STL-10, Caltech-256, Places-365 and CeleBA) to verify whether it can transform clean data into its corresponding adversarial examples. The results are listed in Table \ref{tab:cross}, which illustrates that even on the dataset not involved in the diffusion model training, \Name can also achieve a 90+\% (in some cases, even 100\%) attack success rate. This phenomenon strongly supports our proposition that adversarial examples can be transformed from normal examples. Furthermore, the evasion attack success rate of various query counts is illustrated in Figure~\ref{fig:cross}. As we can see, \Name can achieve a good attack success rate even in relatively fewer queries. Taking Places365 as an example, it can obtain 98\% $\sim$ 99.6\% attack success rate when the noise budget is set as $\epsilon = 8$ and 99.7\% $\sim$ 100\% when $\epsilon = 16$.
	
	\subsubsection{\textbf{Model-independent Attack.}}
\begin{figure}
		\centering 
		\includegraphics[width=0.33\textwidth]{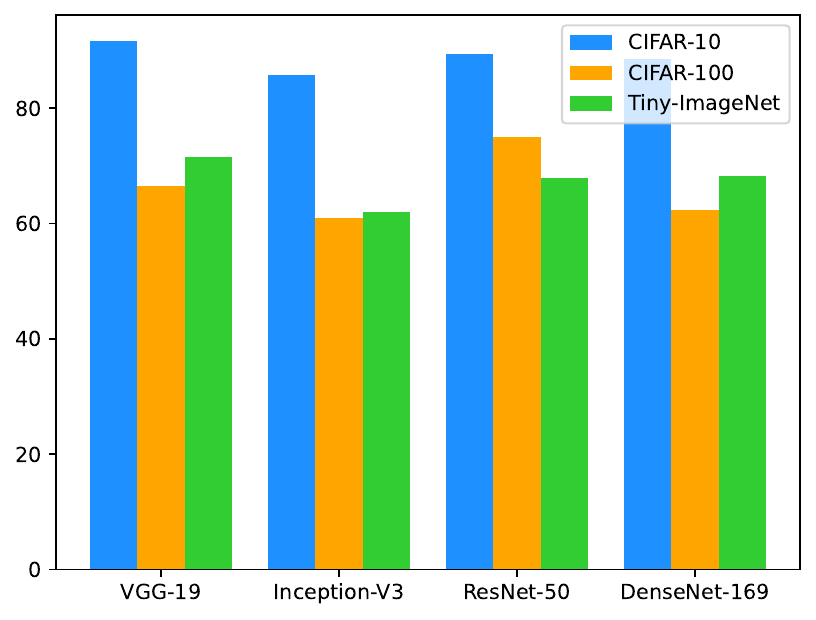}
		\setlength{\abovecaptionskip}{3pt}
		\setlength{\belowcaptionskip}{-20pt}
		\caption{Model-independent attack.}
		\label{fig:mo-in}
	\end{figure}

	Existing black-box attack methods can only generate adversarial samples for a specific victim model. Our \Name is not restricted by this requirement. It can synthesize adversarial examples by performing a conditional sampling manner, and we call it the model-independent attack. Specifically, in this situation, we don't know what the victim model is but just do the conditional sampling once for the specific dataset, and then verify whether these sampled examples are adversarial or not. Figure~\ref{fig:mo-in} shows that the average success rate of such attack on CIFAR-10 is higher than 80\% over different victim models. It can also achieve a 60+\% average attack success rate on CIFAR-100 and Tiny-ImageNet. This phenomenon demonstrates that even in model-independent attack scenarios, \Name can still generate adversarial examples and achieve good attack effects on different models. Furthermore, it illustrates the high adaptability of \Name in model-independent black-box scenarios.
	
	\subsection{Transfer Attack Effectiveness}
	Recall that the transferability of adversarial examples is crucial to carry out transfer attacks, especially for the black-box model deployed in the real world. Therefore following the previous works \cite{nips/DolatabadiEL20,aaai/ZhaoCWL20}, we examine the transferability of the generated AE for each of the attack methods in Table. \ref{tab:trans}. We randomly sample 1000 images from CIFAR-10, CIFAR-100 and Tiny-ImageNet datasets, and generate AEs against on ResNet-50 model. Then we transfer these AEs to attack the VGG-19, Inception V3 and DenseNet-169. As seen, the generated AE by \Name transfer to other models easier than other attacks. This observation precisely matches our intuition about the mechanics of \Name. More specifically, we know that in \Name the model is learning a transformation between a benign image and its adversarial counterpart. Comparing to other queries and optimization attacks, which calculate specific perturbations for each sample, \Name learns to use the transformation to build AEs. Thus, \Name tends to generate AEs with high transferability.
	
	\begin{table}
		\caption{Transfer attack on different models.}
		\label{tab:trans}
		\centering
		\small
		\renewcommand{\arraystretch}{1.1}
		\setlength\tabcolsep{1.2pt}
		\begin{tabular}{c|c|ccc|ccc}
			\hline
			\multirow{2}{*}{}                               & \multirow{2}{*}{Methods} & \multicolumn{3}{c|}{$\epsilon$=8} & \multicolumn{3}{c}{$\epsilon$=16}                                                                     \\
			&                          & VGG                               & Inception                         & DenseNet       & VGG            & Inception      & DenseNet       \\
			\hline
			\multirow{10}{*}{\rotatebox{90}{CIFAR-10}}      & AdvFlow                  & 21.58                             & 9.82                              & 10.78          & 15.75          & 9.43           & 11.89          \\
			& RayS                     & 14.74                             & 7.38                              & 12.74          & 23.53          & 9.76           & 15.00          \\
			& bayes\_attack            & 24.24                             & 34.38                             & 28.01          & 51.44          & 47.54          & 59.10          \\
			& TA                       & 21.15                             & 8.77                              & 19.30          & 23.05          & 26.47          & 24.91          \\
			& NPAttack                 & 12.05                             & 18.84                             & 19.08          & 21.94          & 30.03          & 29.49          \\
			& ODS                      & 50.86                             & 45.94                             & 53.18          & 49.12          & 43.61          & 50.22          \\
			& GFCS                     & 55.92                             & 49.19                             & 55.33          & 54.76          & 51.03          & 57.45          \\
			& CG-Attack                & 57.64                             & 48.28                             & 56.18          & 61.93          & 58.67          & 62.79          \\
			& MCG-Attack               & 53.73                             & 46.91                             & 53.17          & 58.62          & 52.96          & 61.45          \\
			& \Name                    & \textbf{82.43}                    & \textbf{76.06}                    & \textbf{87.13} & \textbf{98.06} & \textbf{95.21} & \textbf{98.51} \\
			\hline
			\multirow{10}{*}{\rotatebox{90}{CIFAR-100}}     & AdvFlow                  & 12.74                             & 5.59                              & 15.30          & 14.93          & 11.48          & 12.83          \\
			& RayS                     & 23.20                             & 9.94                              & 13.90          & 28.08          & 13.52          & 18.09          \\
			& bayes\_attack            & 51.39                             & 52.40                             & 40.04          & 79.86          & 79.13          & 71.89          \\
			& TA                       & 22.61                             & 26.54                             & 24.93          & 45.12          & 39.01          & 30.24          \\
			& NPAttack                 & 39.81                             & 41.38                             & 35.37          & 60.00          & 58.84          & 54.07          \\
			& ODS                      & 50.78                             & 39.65                             & 41.76          & 54.49          & 37.96          & 37.54          \\
			& GFCS                     & 68.57                             & 48.51                             & 52.17          & 65.83          & 49.53          & 53.24          \\
			& CG-Attack                & 61.73                             & 55.49                             & 59.47          & 62.94          & 58.62          & 60.49          \\
			& MCG-Attack               & 58.61                             & 53.97                             & 58.63          & 58.46          & 53.61          & 56.37          \\
			& \Name                    & \textbf{71.74}                    & \textbf{60.96}                    & \textbf{68.55} & \textbf{89.57} & \textbf{82.12} & \textbf{87.71} \\
			\hline
			\multirow{10}{*}{\rotatebox{90}{Tiny-ImageNet}} & AdvFlow                  & 8.82                              & 5.75                              & 12.03          & 8.56           & 7.06           & 12.37          \\
			& RayS                     & 15.97                             & 12.72                             & 18.49          & 23.80          & 20.70          & 20.11          \\
			& bayes\_attack            & 31.84                             & 30.41                             & 46.27          & 63.25          & 64.43          & 72.12          \\
			& TA                       & 18.72                             & 24.40                             & 26.98          & 35.37          & 28.72          & 35.80          \\
			& NPAttack                 & 12.78                             & 13.64                             & 18.53          & 16.25          & 17.38          & 25.28          \\
			& ODS                      & 44.49                             & 31.58                             & 42.20          & 47.04          & 27.89          & 42.16          \\
			& GFCS                     & 63.27                             & 43.28                             & 57.32          & 64.60          & 47.37          & 60.60          \\
			& CG-Attack                & 58.61                             & 46.94                             & 57.37          & 60.98          & 52.61          & 59.16          \\
			& MCG-Attack               & 57.92                             & 44.28                             & 54.37          & 61.94          & 53.63          & 58.61          \\
			& \Name                    & \textbf{76.36}                    & \textbf{68.22}                    & \textbf{86.46} & \textbf{90.33} & \textbf{84.31} & \textbf{95.82} \\
			\hline
		\end{tabular}
		
	\end{table}

	\subsection{Ablation Study}
	\subsubsection{\textbf{Scheduling \& Steps.}} Although the typical training and sampling steps of DDPM are $T=1000$, previous work \cite{siggraph/SahariaCCLHSF022} shows that such number of steps for the diffusion model can be inconsistent. Here, we aim to explore the effect of the number of sampling steps $T$ on the final attack performance without other acceleration schemes \cite{iclr/Liu0LZ22, cvpr/RombachBLEO22}. The victim model is ResNet-50, the noise budget is set as $\epsilon=8$ and $\epsilon=16$, and the maximal number of queries is set as $Q=10$.
	
\begin{figure}[t]
		\centering
		\setlength{\abovecaptionskip}{0pt}
		\setlength{\belowcaptionskip}{-6pt}
		\includegraphics[width=0.48\textwidth]{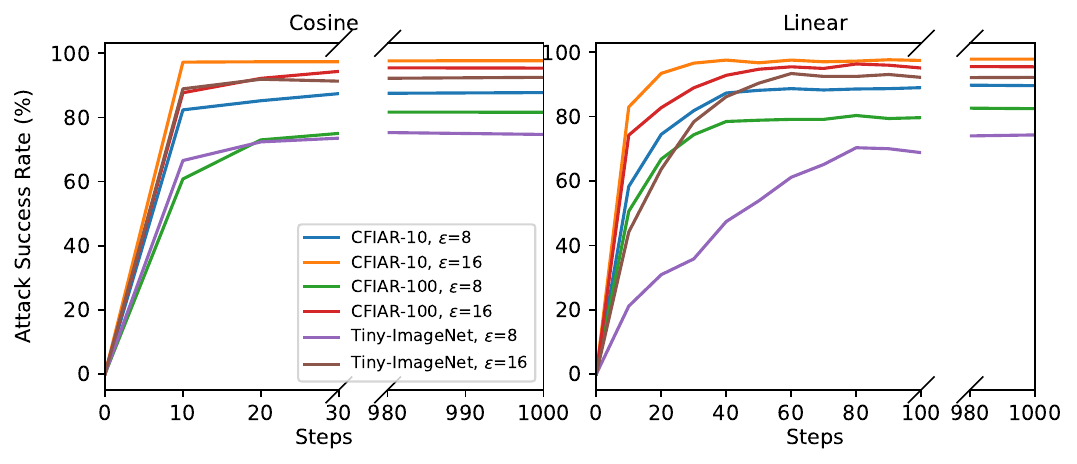}
		\caption{The attack performance v.s. different sample schedules under multiple sample steps.}
		\label{fig:steps}
	\end{figure}

	As shown in Figure~\ref{fig:steps}, the obtained attack success rate always fluctuates regardless of using cosine or linear sampling. Compared with liner sampling, cosine sampling can achieve a higher attack success rate with fewer sampling steps. Especially when the number of sampling steps is small, the attack success rate of linear sampling is relatively lower. For example, when the number of sampling steps is $t=10$, the attack success rate of linear sampling is around 40\%-60\%, while the cosine sampling is 80\%-90\%. Note that for each sampling schedule, the final attack success rate is roughly the same as the number of sampling steps increases. To obtain more effective and efficient attack results, we set the sampling strategies in the attack process as follows: the sampling schedule is cosine and the number of steps is $t=50$. By doing this, the attack effectiveness is significantly promoted owning to a smaller sample step $t$.
	
	\subsubsection{\textbf{Comparisons with Generative Attacks}} 
\begin{figure}[t]
		\centering
		\setlength{\abovecaptionskip}{0pt}
		\setlength{\belowcaptionskip}{-10pt}
		\includegraphics[width=0.48\textwidth]{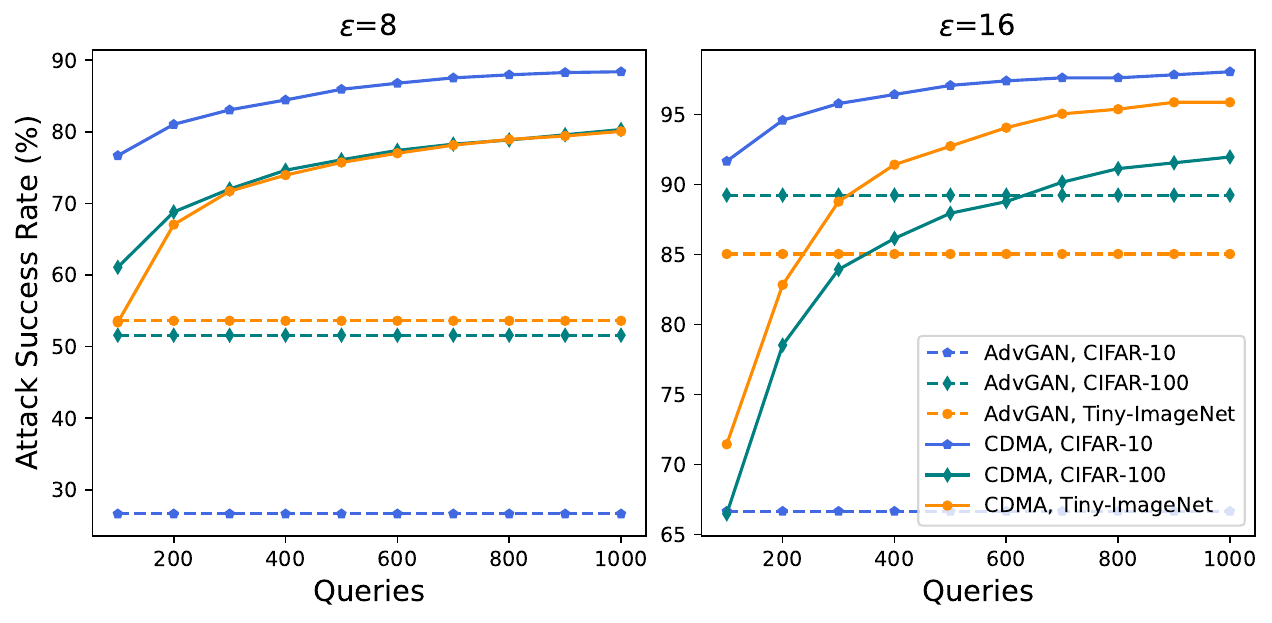}
		\caption{Queries vs. ASR of AdvGAN and CDMA on three different datasets.}
		\label{fig:generative_model}
	\end{figure}

Existing generative-based attacks usually use GAN to generate adversarial perturbations. We choose the most representative one, AdvGAN, among these methods to compare with our CDMA. As the experimental results are illustrated in Figure~\ref{fig:generative_model}, we can find that although the attack success rate of our method is lower than AdvGAN in some cases, as the number of queries increases, the attack success rate of our method will increase with the number of queries, on the contrary, AdvGAN will not, which thoroughly verifies our assertion that the AEs generated by our \Name can generate diversiform adversarial examples, even for the same clean example $x$.
	
	\section{Conclusions}
	\label{Sec:conclusion}
	In this work, we find that adversarial examples are a particular form of benign examples, i.e., these two types of samples come from two distinct but adjacent distributions that can be transformed from each other with a perfect converter. Based on this observation, we propose a novel hard-label black-box attack, \Name, which builds a converter to transform clean data to its corresponding adversarial counterpart. Specifically, we leverage a diffusion model to formulate the data converter and synthesize adversarial examples by conditioning on clean images to improve the query efficiency significantly. Extensive experiments demonstrate that \Name achieves a much higher attack success rate within 1,000 queries and needs fewer queries to achieve the attack results, even in the targeted attack setting. Besides, most adversarial examples generated by \Name can escape from mainstream defense strategies and maintain high robustness. Furthermore, \Name can generate adversarial examples that are well transferred to different victim models or datasets. 
	
	
	\textbf{Acknowledgements} 
	This work is supported in part by Yunnan Province Education Department Foundation under Grant No.2022j0008, in part by the National Natural Science Foundation of China under Grant 62162067 and 62101480, Research and Application of Object Detection based on Artificial Intelligence, in part by the Yunnan Province expert workstations under Grant 202205AF150145.
	
	\bibliographystyle{IEEEtran}
	\bibliography{ieee}

\begin{thebibliography}{10}
\providecommand{\url}[1]{#1}
\csname url@samestyle\endcsname
\providecommand{\newblock}{\relax}
\providecommand{\bibinfo}[2]{#2}
\providecommand{\BIBentrySTDinterwordspacing}{\spaceskip=0pt\relax}
\providecommand{\BIBentryALTinterwordstretchfactor}{4}
\providecommand{\BIBentryALTinterwordspacing}{\spaceskip=\fontdimen2\font plus
\BIBentryALTinterwordstretchfactor\fontdimen3\font minus
  \fontdimen4\font\relax}
\providecommand{\BIBforeignlanguage}[2]{{%
\expandafter\ifx\csname l@#1\endcsname\relax
\typeout{** WARNING: IEEEtran.bst: No hyphenation pattern has been}%
\typeout{** loaded for the language `#1'. Using the pattern for}%
\typeout{** the default language instead.}%
\else
\language=\csname l@#1\endcsname
\fi
#2}}
\providecommand{\BIBdecl}{\relax}
\BIBdecl

\bibitem{pr/ChenYTX22}
J.~Chen, L.~Yang, L.~Tan, and R.~Xu, ``Orthogonal channel attention-based
  multi-task learning for multi-view facial expression recognition,''
  \emph{Pattern Recognition}, vol. 129, p. 108753, 2022.

\bibitem{eccv/LiZOQ22}
G.~Li, Y.~Zhang, D.~Ouyang, and X.~Qu, ``An improved lightweight network based
  on yolov5s for object detection in autonomous driving,'' in \emph{ECCV}, vol.
  13801, 2022, pp. 585--601.

\bibitem{isci/ZhangHXW21}
X.~Zhang, Y.~Han, W.~Xu, and Q.~Wang, ``{HOBA:} {A} novel feature engineering
  methodology for credit card fraud detection with a deep learning
  architecture,'' \emph{Information Sciences}, vol. 557, pp. 302--316, 2021.

\bibitem{sp/Carlini017}
N.~Carlini and D.~A. Wagner, ``Towards evaluating the robustness of neural
  networks,'' in \emph{S\&P}, 2017.

\bibitem{make/CombeyLFH20}
T.~Combey, A.~Loison, M.~Faucher, and H.~Hajri, ``Probabilistic jacobian-based
  saliency maps attacks,'' \emph{Machine Learning and Knowledge Extraction},
  vol.~2, no.~4, pp. 558--578, 2020.

\bibitem{corr/GoodfellowSS14}
I.~J. Goodfellow, J.~Shlens, and C.~Szegedy, ``Explaining and harnessing
  adversarial examples,'' in \emph{ICLR}, 2015.

\bibitem{kdd/ChenG20}
J.~Chen and Q.~Gu, ``Rays: {A} ray searching method for hard-label adversarial
  attack,'' in \emph{KDD}, 2020, pp. 1739--1747.

\bibitem{nips/DolatabadiEL20}
H.~M. Dolatabadi, S.~M. Erfani, and C.~Leckie, ``Advflow: Inconspicuous
  black-box adversarial attacks using normalizing flows,'' in \emph{NeurIPS},
  2020.

\bibitem{kdd/ShuklaSWK21}
S.~N. Shukla, A.~K. Sahu, D.~Willmott, and J.~Z. Kolter, ``Simple and efficient
  hard label black-box adversarial attacks in low query budget regimes,'' in
  \emph{KDD}, 2021, pp. 1461--1469.

\bibitem{eccv/WangZTGHLL22}
X.~Wang, Z.~Zhang, K.~Tong, D.~Gong, K.~He, Z.~Li, and W.~Liu, ``Triangle
  attack: {A} query-efficient decision-based adversarial attack,'' in
  \emph{ECCV}, vol. 13665, 2022, pp. 156--174.

\bibitem{pr/BaiWZJX23}
Y.~Bai, Y.~Wang, Y.~Zeng, Y.~Jiang, and S.~Xia, ``Query efficient black-box
  adversarial attack on deep neural networks,'' \emph{Pattern Recognition},
  vol. 133, p. 109037, 2023.

\bibitem{iclr/BrendelRB18}
W.~Brendel, J.~Rauber, and M.~Bethge, ``Decision-based adversarial attacks:
  Reliable attacks against black-box machine learning models,'' in \emph{ICLR},
  2018.

\bibitem{iclr/IlyasEM19}
A.~Ilyas, L.~Engstrom, and A.~Madry, ``Prior convictions: Black-box adversarial
  attacks with bandits and priors,'' in \emph{ICLR}, 2019.

\bibitem{mm/ShahRKR21}
S.~B. Shah, P.~Raval, H.~Khakhi, and M.~S. Raval, ``Frequency centric defense
  mechanisms against adversarial examples,'' in \emph{MM}, 2021, pp. 62--67.

\bibitem{icml/NieGHXVA22}
W.~Nie, B.~Guo, Y.~Huang, C.~Xiao, A.~Vahdat, and A.~Anandkumar, ``Diffusion
  models for adversarial purification,'' in \emph{ICML}, vol. 162, 2022, pp.
  16\,805--16\,827.

\bibitem{corr/abs-2205-14969}
J.~Wang, Z.~Lyu, D.~Lin, B.~Dai, and H.~Fu, ``Guided diffusion model for
  adversarial purification,'' \emph{CoRR}, vol. abs/2205.14969, 2022.

\bibitem{ccs/PapernotMGJCS17}
N.~Papernot, P.~D. McDaniel, I.~J. Goodfellow, S.~Jha, Z.~B. Celik, and
  A.~Swami, ``Practical black-box attacks against machine learning,'' in
  \emph{Asia@CCS}, 2017, pp. 506--519.

\bibitem{iclr/Huang020}
Z.~Huang and T.~Zhang, ``Black-box adversarial attack with transferable
  model-based embedding,'' in \emph{ICLR}, 2020.

\bibitem{nips/Tashiro0E20}
Y.~Tashiro, Y.~Song, and S.~Ermon, ``Diversity can be transferred: Output
  diversification for white- and black-box attacks,'' in \emph{NeurIPS}, 2020.

\bibitem{iclr/LordMB22}
N.~A. Lord, R.~M{\"{u}}ller, and L.~Bertinetto, ``Attacking deep networks with
  surrogate-based adversarial black-box methods is easy,'' in \emph{ICLR},
  2022.

\bibitem{cvpr/FengWFL0X22}
Y.~Feng, B.~Wu, Y.~Fan, L.~Liu, Z.~Li, and S.~Xia, ``Boosting black-box attack
  with partially transferred conditional adversarial distribution,'' in
  \emph{CVPR}, 2022, pp. 15\,074--15\,083.

\bibitem{ccs/ChenZSYH17}
P.~Chen, H.~Zhang, Y.~Sharma, J.~Yi, and C.~Hsieh, ``{ZOO:} zeroth order
  optimization based black-box attacks to deep neural networks without training
  substitute models,'' in \emph{AISec@CCS}, 2017, pp. 15--26.

\bibitem{nips/HoJA20}
J.~Ho, A.~Jain, and P.~Abbeel, ``Denoising diffusion probabilistic models,'' in
  \emph{NeurIPS}, 2020.

\bibitem{miccai/RonnebergerFB15}
O.~Ronneberger, P.~Fischer, and T.~Brox, ``U-net: Convolutional networks for
  biomedical image segmentation,'' in \emph{MICCAI}, vol. 9351, 2015, pp.
  234--241.

\bibitem{iclr/MadryMSTV18}
A.~Madry, A.~Makelov, L.~Schmidt, D.~Tsipras, and A.~Vladu, ``Towards deep
  learning models resistant to adversarial attacks,'' in \emph{ICLR}, 2018.

\bibitem{siggraph/SahariaCCLHSF022}
C.~Saharia, W.~Chan, H.~Chang, C.~A. Lee, J.~Ho, T.~Salimans, D.~J. Fleet, and
  M.~Norouzi, ``Palette: Image-to-image diffusion models,'' in \emph{SIGGRAPH},
  2022, pp. 15:1--15:10.

\bibitem{iccv/ChoiKJGY21}
J.~Choi, S.~Kim, Y.~Jeong, Y.~Gwon, and S.~Yoon, ``{ILVR:} conditioning method
  for denoising diffusion probabilistic models,'' in \emph{ICCV}, 2021, pp.
  14\,347--14\,356.

\bibitem{cifar-10}
A.~Krizhevsky and G.~Hinton, ``Learning multiple layers of features from tiny
  images,'' \emph{Computer Science Department, University of Toronto, Tech.
  Rep}, vol.~1, 01 2009.

\bibitem{cvpr/DengDSLL009}
J.~Deng, W.~Dong, R.~Socher, L.~Li, K.~Li, and L.~Fei{-}Fei, ``Imagenet: {A}
  large-scale hierarchical image database,'' in \emph{CVPR}, 2009, pp.
  248--255.

\bibitem{corr/SimonyanZ14a}
K.~Simonyan and A.~Zisserman, ``Very deep convolutional networks for
  large-scale image recognition,'' in \emph{ICLR}, 2015.

\bibitem{cvpr/SzegedyVISW16}
C.~Szegedy, V.~Vanhoucke, S.~Ioffe, J.~Shlens, and Z.~Wojna, ``Rethinking the
  inception architecture for computer vision,'' in \emph{CVPR}, 2016, pp.
  2818--2826.

\bibitem{aaai/SzegedyIVA17}
C.~Szegedy, S.~Ioffe, V.~Vanhoucke, and A.~A. Alemi, ``Inception-v4,
  inception-resnet and the impact of residual connections on learning,'' in
  \emph{AAAI}, 2017, pp. 4278--4284.

\bibitem{cvpr/HeZRS16}
K.~He, X.~Zhang, S.~Ren, and J.~Sun, ``Deep residual learning for image
  recognition,'' in \emph{CVPR}, 2016, pp. 770--778.

\bibitem{cvpr/HuangLMW17}
G.~Huang, Z.~Liu, L.~van~der Maaten, and K.~Q. Weinberger, ``Densely connected
  convolutional networks,'' in \emph{CVPR}, 2017, pp. 2261--2269.

\bibitem{corr/abs-2301-00364}
F.~Yin, Y.~Zhang, B.~Wu, Y.~Feng, J.~Zhang, Y.~Fan, and Y.~Yang,
  ``Generalizable black-box adversarial attack with meta learning,''
  \emph{CoRR}, vol. abs/2301.00364, 2023.

\bibitem{shin2017jpeg}
R.~Shin and D.~Song, ``Jpeg-resistant adversarial images,'' in \emph{NeurIPS},
  vol.~1, 2017, p.~8.

\bibitem{nips/Naseer0KKP19}
M.~Naseer, S.~H. Khan, M.~H. Khan, F.~S. Khan, and F.~Porikli, ``Cross-domain
  transferability of adversarial perturbations,'' in \emph{NeurIPS}, 2019, pp.
  12\,885--12\,895.

\bibitem{cvpr/PrakashMGDS18}
A.~Prakash, N.~Moran, S.~Garber, A.~DiLillo, and J.~A. Storer, ``Deflecting
  adversarial attacks with pixel deflection,'' in \emph{CVPR}, 2018, pp.
  8571--8580.

\bibitem{corr/abs-1810-00953}
``Improved robustness to adversarial examples using lipschitz regularization of
  the loss,'' \emph{CoRR}, vol. abs/1810.00953, 2018.

\bibitem{ndss/Xu0Q18}
W.~Xu, D.~Evans, and Y.~Qi, ``Feature squeezing: Detecting adversarial examples
  in deep neural networks,'' in \emph{NDSS}, 2018.

\bibitem{jmlr/CoatesNL11}
A.~Coates, A.~Y. Ng, and H.~Lee, ``An analysis of single-layer networks in
  unsupervised feature learning,'' in \emph{AISTATS}, ser. {JMLR} Proceedings,
  vol.~15, 2011, pp. 215--223.

\bibitem{cviu/Fei-FeiFP07}
L.~Fei{-}Fei, R.~Fergus, and P.~Perona, ``Learning generative visual models
  from few training examples: An incremental bayesian approach tested on 101
  object categories,'' \emph{computer vision and image understanding}, vol.
  106, no.~1, pp. 59--70, 2007.

\bibitem{zhou2017places}
B.~Zhou, A.~Lapedriza, A.~Khosla, A.~Oliva, and A.~Torralba, ``Places: A 10
  million image database for scene recognition,'' \emph{IEEE Transactions on
  Pattern Analysis and Machine Intelligence}, 2017.

\bibitem{iccv/LiuLWT15}
Z.~Liu, P.~Luo, X.~Wang, and X.~Tang, ``Deep learning face attributes in the
  wild,'' in \emph{ICCV}, 2015, pp. 3730--3738.

\bibitem{aaai/ZhaoCWL20}
P.~Zhao, P.~Chen, S.~Wang, and X.~Lin, ``Towards query-efficient black-box
  adversary with zeroth-order natural gradient descent,'' in \emph{AAAI}, 2020,
  pp. 6909--6916.

\bibitem{iclr/Liu0LZ22}
L.~Liu, Y.~Ren, Z.~Lin, and Z.~Zhao, ``Pseudo numerical methods for diffusion
  models on manifolds,'' in \emph{ICLR}, 2022.

\bibitem{cvpr/RombachBLEO22}
R.~Rombach, A.~Blattmann, D.~Lorenz, P.~Esser, and B.~Ommer, ``High-resolution
  image synthesis with latent diffusion models,'' in \emph{CVPR}, 2022, pp.
  10\,674--10\,685.

\end{thebibliography}

\end{document}